\def\paperlanguage{}
\pdfoutput=1 

\documentclass[letterpaper, 10 pt, conference]{ieeeconf}  

\usepackage{bm}
\usepackage{cite}
\usepackage{flushend}
\include{preamble}

\IEEEoverridecommandlockouts                              

\title{\LARGE \textbf
  {
    \switchlanguage%
    {%
      Applications of Stretch Reflex\\for the Upper Limb of Musculoskeletal Humanoids:\\Protective Behavior, Postural Stability, and Active Induction
    }%
    {%
      筋骨格ヒューマノイドの上肢における伸張反射のアプリケーション
    }%
  }
}

\author{Kento Kawaharazuka$^{1}$, Yuya Koga$^{1}$, Kei Tsuzuki$^{1}$, Moritaka Onitsuka$^{1}$, Yuki Asano$^{1}$, Kei Okada$^{1}$,\\Koji Kawasaki$^{2}$, and Masayuki Inaba$^{1}$
  \thanks{$^{1}$ The authors are with the Department of Mechano-Informatics, Graduate School of Information Science and Technology, The University of Tokyo, 7-3-1 Hongo, Bunkyo-ku, Tokyo, 113-8656, Japan.
    {\texttt\small [kawaharazuka, koga, tsuzuki, onitsuka, asano, k-okada, inaba]@jsk.t.u-tokyo.ac.jp}
  }
  \thanks{$^{2}$ The author is associated with TOYOTA MOTOR CORPORATION.
    {\texttt\small koji\_kawasaki@mail.toyota.co.jp}
  }
}
\begin{document}

\maketitle
\thispagestyle{empty}
\pagestyle{empty}

\begin{abstract}
  \switchlanguage%
  {%
    The musculoskeletal humanoid has various biomimetic benefits, and it is important that we can embed and evaluate human reflexes in the actual robot.
    Although stretch reflex has been implemented in lower limbs of musculoskeletal humanoids, we apply it to the upper limb to discover its useful applications.
    We consider the implementation of stretch reflex in the actual robot, its active/passive applications, and the change in behavior according to the difference of parameters.
  }%
  {%
    筋骨格ヒューマノイドには様々な生物規範型の利点が存在するが, その中でも, 人間の反射をそのまま実装して組み込み, ロボット上で評価を行える点は重要である.
    今回扱う伸張反射は歩行の安定化のために下半身に多く実装されてきたが, 本研究では筋骨格ヒューマノイドの上肢にこれを適用し, そのアプリケーションを見出す.
    伸張反射の実ロボットにおける実装, 伸張反射の能動的/受動的利用や, パラメータの違いによる振る舞いの変化等について考察し, 実ロボットへの適用についての有用性を示す.
  }%
\end{abstract}

\section{INTRODUCTION}\label{sec:introduction}
\switchlanguage%
{%
  The musculoskeletal humanoid \cite{nakanishi2013design, wittmeier2013toward, jantsch2013anthrob, kawaharazuka2019musashi} has various biomimetic benefits such as redundant muscle arrangement \cite{kawaharazuka2019longtime}, ball joints without singular points, and the flexible spine \cite{mizuuchi2004kenta} and fingers \cite{makino2018hand}.
  Also, the fact that we can directly apply human control schemes to such human-like structures is important.
  Especially, human reflexes are important for human beings to survive, and it is considered to be also useful for musculoskeletal humanoids.

  There are various kinds of human reflexes, and the most basic reflexes are stretch reflex, goldi tendon reflex, and reciprocal innervation.
  We introduce examples of applying these reflexes to the musculoskeletal structure of a simulation or the actual robot.
  Stretch reflex is used to increase the stability of jumping \cite{liu2018stretch, shimizu2012jumping} and walking \cite{geyer2010reflex}.
  Reciprocal innervation is effective to permit model error and conduct a wide range of limb motions \cite{kawaharazuka2017antagonist}.
  \cite{folgheraiter2004reflex} has embedded all the three reflexes into the fingers and showed the possibility of moving safely, especially by goldi tendon reflex.
  \cite{endo1994reflex} has verified that these reflexes can stabilize the whole control system.
  Also, \cite{marques2013reflex} has discovered the self-organization of reflexive behaviors from spontaneous motor activities using a Hebb learning rule and dynamics simulation.

  From these studies, we can see that stretch reflex is useful for postural stability of lower limbs, goldi tendon reflex is useful for safe motions, and reciprocal innervation is useful for the reduction of internal muscle tensions and efficient movements.
  However, stretch reflex has been effectively used for only lower limbs, and we have not found effective applications for the upper limbs of the actual musculoskeletal humanoids.
  Therefore, we handle stretch reflex of the upper limbs in this study.
  We apply it to the musculoskeletal humanoid Musashi \cite{kawaharazuka2019musashi}, and construct hypotheses about its effective applications.
  We verify the hypotheses using the actual robot.

  This paper is organized as follows.
  In \secref{sec:musculoskeletal-humanoids}, we will explain the basic musculoskeletal structure and the human stretch reflex.
  In \secref{sec:proposed}, we will consider the implementation of stretch reflex and the classification of its applications.
  In \secref{sec:experiments}, we will conduct four experiments from the classification.
  Finally, we will discuss our experimental results and state the conclusion.
}%
{%
  筋骨格ヒューマノイド\cite{nakanishi2013design, wittmeier2013toward, jantsch2013anthrob, kawaharazuka2019musashi}は冗長な筋配置\cite{kawaharazuka2019longtime}や球関節, 柔軟な背骨\cite{mizuuchi2004kenta}や指\cite{makino2018hand}等, 人間と同様の生物規範型の構造を多く有する.
  また, 人間と同様な構造には, 人間と同様な制御を直接適用できる点も重要である.
  特に, 人間の反射制御は生命活動において重要であり, 実ロボットにおいても役に立つと考えられている.

  人間の反射には様々な種類が存在するが, その中でも最も基本的なのは, 伸張反射, ゴルジ腱反射, 相反性神経支配であろう.
  これらをシミュレーションまたは実機の筋骨格構造に適用した例を紹介する.
  伸張反射はジャンプ時\cite{liu2018stretch, shimizu2012jumping}, または歩行時\cite{geyer2010reflex}の安定性を増すことができることが示されている.
  相反性神経支配はモデル誤差を許した広可動域動作\cite{kawaharazuka2017antagonist}に有効であることが示されている.
  手の指に伸張反射やゴルジ腱反射, 相反性神経支配を組み込むことで, 特にゴルジ腱反射によって身体を安全に動作させることができることが示されている\cite{folgheraiter2004reflex}.
  制御理論的に, これらの反射制御がシステムを安定化し, interactionを削減することができることが示されている\cite{endo1994reflex}.
  また, Hebb則による学習を用いることで, シミュレーション上ではあるが, これらの反射が詳細な筋骨格モデルにおいて自己組織化することが示されている\cite{marques2013reflex}.

  これらから, 伸張反射は下半身の安定的な姿勢保持, ゴルジ腱反射は身体を安全に保つこと, 相反性神経支配は拮抗を減らし安全かつ効率的に動作することに有用であることがわかる.
  しかし, 伸張反射は主に下半身の制御のみに有用性が見出されており, 特に実機において上肢における有用性・実際のアプリケーションが見出されていない.
  そこで本研究では, 上肢に関する伸張反射について扱う.
  筋骨格ヒューマノイドMusashiに伸張反射を実装し, そのアプリケーションについていくつかの仮説を構築する.
  実機においてそれらの実証実験を試みることで, その有用性を明らかにする.

  本研究は以下のように構成される.
  \secref{sec:musculoskeletal-humanoids}では筋骨格ヒューマノイドの基本的な構造, そして人間の伸張反射について述べる.
  \secref{sec:proposed}では, 伸張反射の実装方法, その応用先の分類について考察する.
  \secref{sec:experiments}では, 先の応用に関する分類から4つの実験を行う.
  最後に, 実験に関する考察を行い, 結論を述べる.
}%


\begin{figure}[t]
  \centering
  \includegraphics[width=0.8\columnwidth]{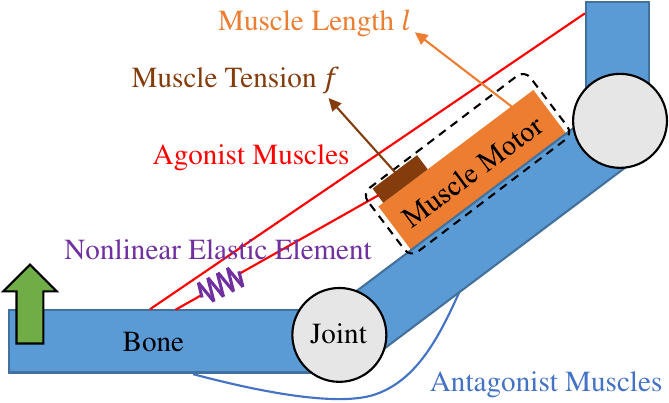}
  \caption{The basic musculoskeletal structure.}
  \label{figure:musculoskeletal-structure}
  \vspace{-3.0ex}
\end{figure}

\section{Musculoskeletal Humanoids and Human Stretch Reflex} \label{sec:musculoskeletal-humanoids}
\subsection{The Basic Structure of the Musculoskeletal Humanoid} \label{subsec:basic-structure}
\switchlanguage%
{%
  In this study, although we handle tendon-driven structures whose muscle wires are wound by electric motors, the same principle as this study can be applied to pneumatic-driven musculoskeletal structures.
  We show the basic musculoskeletal structure in \figref{figure:musculoskeletal-structure}.
  Monoarticular and biarticular muscles are redundantly arranged around joints.
  The muscles contributing in the direction of the movement are agonist muscles, and the muscles contributing in the direction to inhibit the movement are antagonist muscles.
  The abrasion resistant synthetic fiber (e.g. Dyneema) is used as the muscle wires, and it is wound by electric motors.
  Muscle length $l$ can be measured from the encoder attached at the motor, and muscle tension $f$ can be measured using a loadcell or strain gauge.
  A nonlinear elastic element is usually attached to the endpoint of the muscle, in order to enable variable stiffness control.
  Although the joint angle cannot be usually measured, some musculoskeletal humanoids have joint angle sensors for experimental verification or learning \cite{kawaharazuka2019musashi}.
}%
{%
  本研究では, 主に筋が電気モーターによって巻き取られる方式の腱駆動構造を対象とするが, 同様の原理は空気圧型の筋骨格ヒューマノイドにも適用可能であると考える.
  筋骨格ヒューマノイドの基本的な構造を\figref{figure:musculoskeletal-structure}に示す.
  関節の周りに単関節筋や二関節筋が冗長に配置されている.
  動作させる方向に寄与する筋は主動筋, 阻害する方向に寄与する筋は拮抗筋と呼ばれる.
  筋は摩擦に強い合成繊維(例えばダイニーマ)が用いられ, これを電気モータが巻き取ることで駆動する.
  電気モータのエンコーダから筋長$l$, また, ロードセルや歪ゲージを使って筋張力$f$が測定できる.
  筋の終止点には可変剛性制御を可能とする非線形弾性要素が取り付けられている場合が多い.
  関節角度は人間と同様基本的には測定できないが, 検証用や学習用として関節角度センサを有する筋骨格ヒューマノイドも存在する\cite{kawaharazuka2019musashi}.
}%


\subsection{Stretch Reflex in Human Beings} \label{subsec:reflex-control}
\switchlanguage%
{%
  The Human muscle has two sensory receptors of the muscle spindle and goldi tendon organ.
  The muscle spindle is the spindle-shaped organ attached in parallel with the muscle fiber, and can detect muscle length $l$ and muscle velocity $\dot{l}$.
  The goldi tendon organ is arranged at the edge of the muscle in series with the muscle fiber, and can detect muscle tension $f$.
  Group Ia fiber from the muscle spindle is directly connected to the $\alpha$ motor neurons through excitatory connections.
  When the muscle is suddenly stretched, the frequency of impulse from the muscle spindle increases, it excites $\alpha$ motor neurons, and the stretched muscle finally contracts.
  This creates the reflex loop called stretch reflex, and it is a negative feedback system of muscle length.
}%
{%
  人間の反射制御の概要を\figref{figure:reflex-principle}に示す.
  筋紡錘は筋繊維に平行に付着する紡錘形の器官であり、筋の長さ$l$や短縮速度$\dot{l}$を検知する受容器である。
  筋紡錘からはIa群繊維が脊髄の$\alpha$運動ニューロンに直接興奮性接続されており、何らかの原因で急に筋が伸ばされると筋紡錘からのインパルス頻度が増大し、$\alpha$運動ニューロンを興奮させ最終的に伸ばされた筋が収縮する。
  これは伸張反射と呼ばれる反射ループを成し、筋長を出力とする負のフィードバック系を成す。

  腱器官は筋の端に配置されており、筋紡錘が筋繊維と平行に配置されているのに対し、筋繊維と直列に接続する。
  腱器官からはIb群繊維が抑制性介在ニューロンを経て脊髄の$\alpha$運動ニューロンに接続しているため、筋が収縮し筋張力が増すと腱器官からのインパルス頻度が増大し、$\alpha$運動ニューロンの興奮作用が抑制される。
  これは腱反射系とよばれる反射ループを成し、筋張力を出力とする負のフィードバック系を成す。

  筋紡錘からのIa求心性繊維は主動筋の運動ニューロンを興奮させると同時に、抑制性のニューロンを介して拮抗筋の運動ニューロンを抑制するような神経回路が用意されている。
  このような拮抗筋と主動筋の運動ニューロン間の相互作用を相反性神経支配という。

  以下, 本研究ではこの中でも伸張反射を筋骨格ヒューマノイドの実機に実装し, その有用性を確認していく.
}%

\section{Stretch Reflex for Musculoskeletal Humanoids} \label{sec:proposed}
\subsection{Implementation of Stretch Reflex} \label{subsec:implementation}
\begin{figure}[t]
  \centering
  \includegraphics[width=0.4\columnwidth]{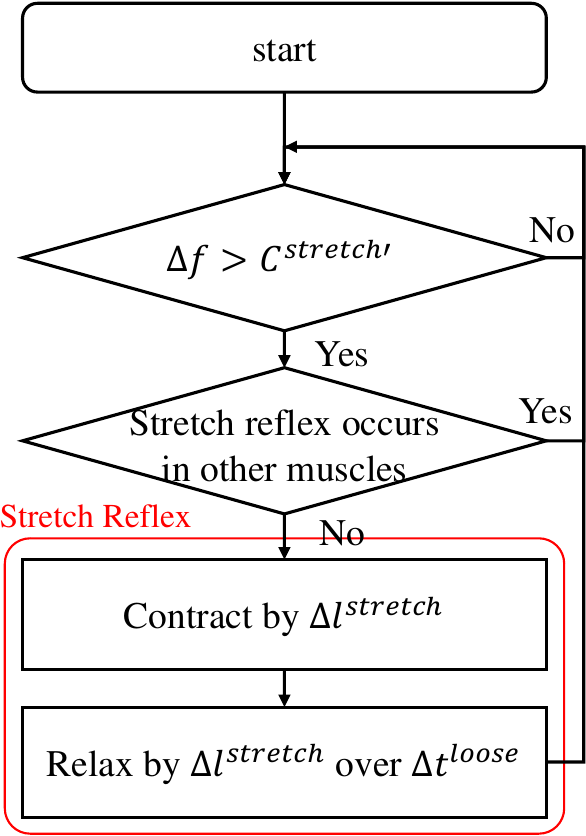}
  \caption{The flow chart of the implemented stretch reflex.}
  \label{figure:reflex-control}
  \vspace{-1.0ex}
\end{figure}

\switchlanguage%
{%
  As explained in \secref{subsec:reflex-control}, stretch reflex contracts the muscle when the muscle length $l$ is stretched quickly.
  Here, we consider a condition of inducing stretch reflex.
  From the definition above, when the difference of muscle lengths (muscle velocity) $\Delta{l}=l_{t+1}-l_{t}$ ($l_{t}$ is the muscle length at the time step $t$) exceeds a threshold $C^{stretch}$, stretch reflex occurs.
  However, as explained in \secref{subsec:basic-structure}, the musculoskeletal humanoids usually have nonlinear elastic elements in the muscles.
  In this case, even when the arms are suddenly pushed, the deformation of the nonlinear elastic elements, which can flexibly react to the impact, is dominant compared to the movement of motors.
  By setting the muscle tension applied to the nonlinear elastic element as $f$ and the relationship between $f$ and its elongation $\Delta{n}$ as $f = {e}^{k\Delta{n}}$ ($k$ is a constant), we can judge whether the muscle is stretched or not, as below,
  \begin{align}
    \Delta{n}_{t+1}-\Delta{n}_{t} &> C^{stretch}\nonumber\\
    \frac{1}{k}\textrm{log}(f_{t+1})-\frac{1}{k}\textrm{log}(f_{t}) &> C^{stretch}\nonumber\\
    f_{t+1}-f_{t} &> C^{stretch'}
  \end{align}
  where $C^{stretch'}=e^{kC^{stretch}}$.
  When the difference of muscle tension $\Delta{f}=f_{t+1}-f_{t}$ exceeds $C^{stretch'}$, stretch reflex occurs.

  Next, we consider the behavior of stretch reflex.
  As its definition, stretch reflex contracts the reference muscle length $l^{ref}$ by a constant value $\Delta{l}^{stretch}$ when the conditions are satisfied.
  After that, $l^{ref}$ gradually loosens by $\Delta{l}^{stretch}$ over $\Delta{t}^{loose}$ seconds.
  Here, to avoid the situation in which the stretch reflex of a certain muscle induces that of other muscles around the muscle, making the antagonistic muscles vibrate coordinately, stretch reflex of other muscles around the contracted muscle should be inhibited.
  Therefore, the flow chart of stretch reflex is as shown in \figref{figure:reflex-control}.
  Because the process of \figref{figure:reflex-control} occurs in parallel with all the muscles, stretch reflex of multiple muscles can occur when the conditions of stretch reflex are satisfied in the multiple muscles at the same time.
  Also, by applying this process independently to limbs that do not induce stretch reflex to each other, even if the timing is off, stretch reflex can occur in each limb (e.g. the left and right arms).

  There are three parameters, $C^{stretch'}$, $\Delta{l}^{stretch}$, and $\Delta{t}^{loose}$, which determine the behavior of stretch reflex.
  We will discuss the difference of behaviors by the change in parameters in the subsequent experimental sections.
}%
{%
  \secref{subsec:reflex-control}に示したように, 伸張反射は筋長$l$が素早く伸ばされたときに, 筋を収縮させるような制御である.
  ここで, 伸張反射が起こる条件について考える.
  先ほどの定義からすると, 一般的には筋長の速度$\Delta{l}=l_{t+1}-l_{t}$ ($l_{t}$は時刻$t$の筋長を表す)が, 閾値$C^{stretch}$よりも大きくなったときに伸張反射が発生する.
  しかし, \secref{subsec:basic-structure}で述べたように, これらのヒューマノイドには非線形弾性要素が含まれていることがほとんどである.
  この場合, 何かの衝撃で急に腕が押されても, モータよりも柔軟で衝撃を受けることが可能な非線形弾性要素の変形が支配的となる.
  非線形弾性要素にかかる筋張力$f$とその伸び$\Delta{n}$の関係を指数関数として$f = {e}^{k\Delta{n}}$ ($k$は定数とする)とすると, 以下のように式を変形することで, 筋が伸ばされたかどうかを判定できる.
  \begin{align}
    \Delta{n}_{t+1}-\Delta{n}_{t} &> C^{stretch}\nonumber\\
    \frac{1}{k}\textrm{log}(f_{t+1})-\frac{1}{k}\textrm{log}(f_{t}) &> C^{stretch}\nonumber\\
    f_{t+1}-f_{t} &> C^{stretch'}
  \end{align}
  ここで, $C^{stretch'}=e^{kC^{stretch}}$とする.
  つまり, 筋張力の前時刻との差分$\Delta{f}=f_{t+1}-f_{t}$が$C^{stretch'}$より大きくなったときに, 伸張反射を発生させれば良い.

  次に, 伸張反射の動作について考える.
  これは定義通り, 条件を満たした瞬間に指令筋長$l^{ref}$をある一定長さ$\Delta{l}^{stretch}$だけ収縮させる.
  その後, $\Delta{t}^{loose}$時間かけて収縮された$\Delta{l}^{stretch}$分の筋を元に戻していく.
  ここで, ある筋の伸張反射により他の筋の伸張反射が誘発されてしまうのを防ぐため, 一つの筋に伸張反射が起きたら, 周辺の筋に関しては伸張反射を起こさないように抑制する.
  ゆえに, 伸張反射のフローチャートは\figref{figure:reflex-control}のようになる.
  なお, \figref{figure:reflex-control}のプロセスは全ての筋で同時並列的に起こるため, 同じタイムステップ時に条件を満たすことができれば, 同時に複数の筋が伸張反射を起こすことがあり得る.
  また, 右腕, 左腕など, 明らかに伸張反射を連鎖的に誘発しない独立した部位に分けてそれぞれこのプロセスを行うことで, タイミングがズレていても, 右腕, 左腕にそれぞれ伸張反射が起こりうる.

  ここで, 本研究における伸張反射の挙動を決めるパラメータは$C^{stretch'}$, $\Delta{l}^{stretch}$, $\Delta{t}^{loose}$の3つである.
  これらのパラメータによる伸張反射の違いは, 後の実験で議論する.
}%

\subsection{Applications of Stretch Reflex}\label{subsec:applications}
\begin{figure}[t]
  \centering
  \includegraphics[width=0.8\columnwidth]{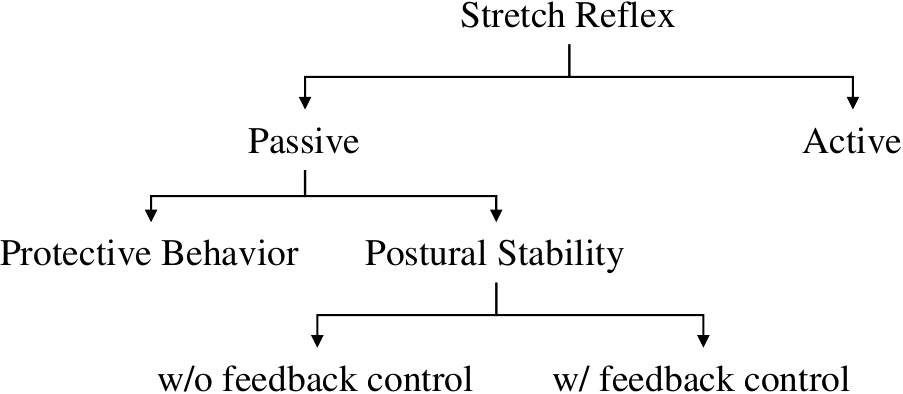}
  \caption{The classification of applications of stretch reflex.}
  \label{figure:reflex-classification}
  \vspace{-3.0ex}
\end{figure}

\switchlanguage%
{%
  We consider useful applications of stretch reflex when it is applied to the upper limbs.
  We show the classification of applications in \figref{figure:reflex-classification}.
  In this study, first, we divide it into passive and active applications.

  The passive applications are the cases in which stretch reflex is passively induced by sudden impact to the upper limbs.
  This is divided into protective behavior and postural stability.
  In protective behavior, it is hypothesized that stretch reflex can keep the posture in the safe range of joint angles when sudden impact occurs around the joint angle limit.
  In postural stability, it is hypothesized that stretch reflex can keep the same posture of the upper limbs before and after the sudden impact.
  Also, we can divide postural stability into the case without any feedback controls (the constant muscle length is kept) and with a feedback control (the joint angle is constantly kept by joint angle feedback control).

  In the active application, it is hypothesized that stretch reflex can exert a faster and larger force than ordinary motions by applying excessive force to the environment and inducing stretch reflex on purpose.
  Weight lifting and jumping are considered to be good examples.

}%
{%
  この伸張反射を上肢に適用したときに, どのようなアプリケーションが考えられるかについて考える.
  その簡単な分類を\figref{figure:reflex-classification}に示す.
  本研究では, それらをまず受動的・能動的という二つに分けた.

  受動的とは, 上肢への突然の衝撃によって伸張反射が発生するような場合である.
  そしてこれは, 危険回避と姿勢保持という2つに分類される.
  危険回避とは, 関節角度リミット近くで衝撃が加わった場合, 伸張反射を入れることで, リミットまで到達せず, 身体を安全な可動範囲に保つことが可能であるという仮説である.
  姿勢保持とは, 突然の衝撃が加わったあと, 伸張反射があることで, 上肢の関節角度を衝撃が加わる前と同じに保つことが可能であるという仮説である.
  また, 姿勢保持にも, 身体に制御が入っていない, つまり一定筋長を保っているような場合と, 身体に制御が入り, 自身の関節角度を一定に保とうとフィードバックが行われている場合という2つに分けて考えることができる.

  能動的とは, 自分から環境に力を加え, 伸張反射を起こすことで, 通常よりも素早く・強い力を出すことができるという仮説である.
  重量挙げやジャンプがそのいい例だと考える.

  本研究ではこれらの仮説のもと, 筋骨格ヒューマノイドの実機に置いて実験を行い, 伸張反射の有効性について検証していく.
}%

\begin{figure}[t]
  \centering
  \includegraphics[width=1.0\columnwidth]{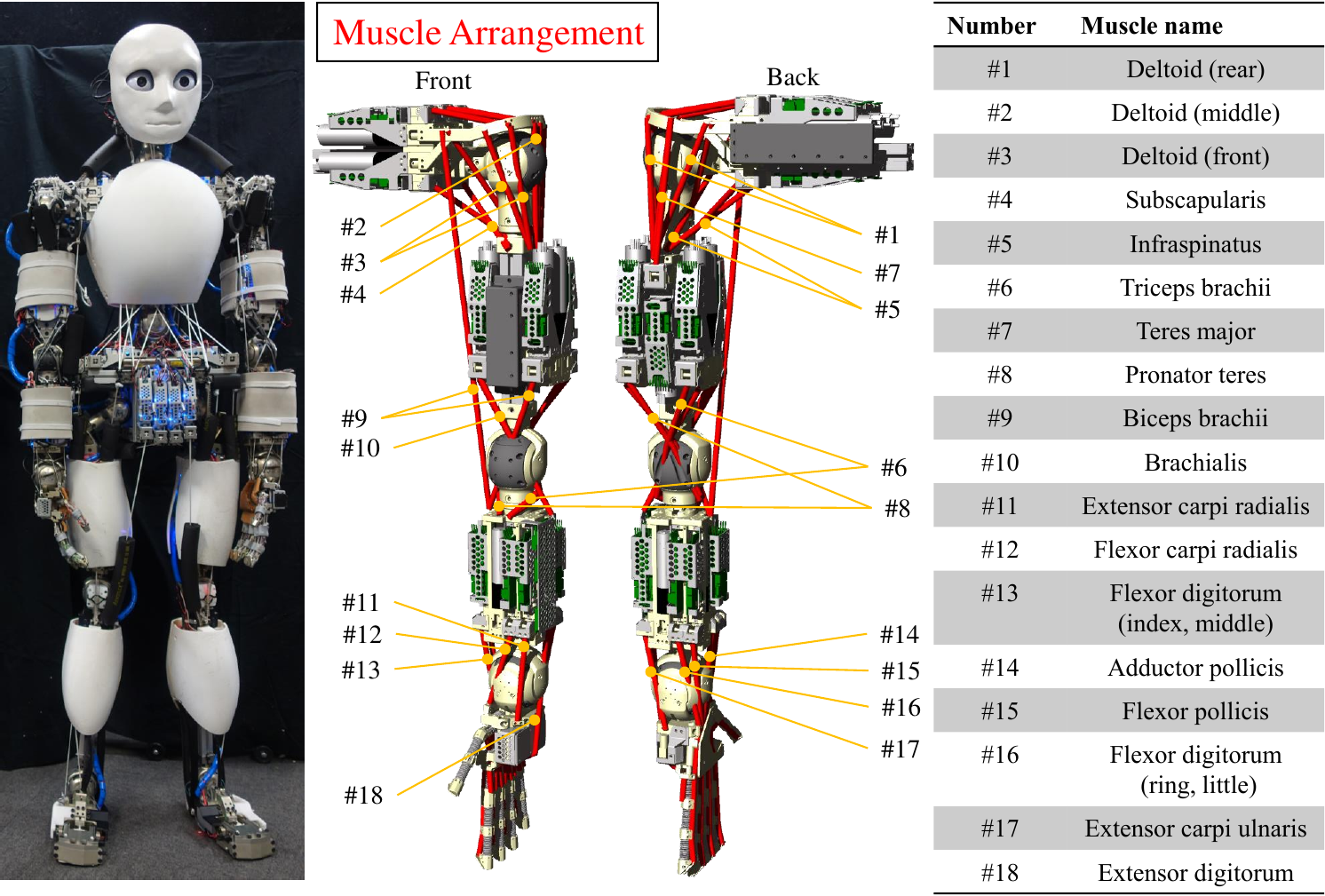}
  \caption{Muscle arrangement of the musculoskeletal humanoid Musashi used in our experiments.}
  \label{figure:experimental-setup}
  \vspace{-1.0ex}
\end{figure}

\section{Experiments} \label{sec:experiments}
\subsection{Experimental Setup}
\switchlanguage%
{%
  We show the musculoskeletal humanoid Musashi \cite{kawaharazuka2019musashi} used in this study and the muscle arrangement of its left arm in \figref{figure:experimental-setup}.
  As in \secref{subsec:basic-structure}, muscle tension and length of Musashi can be measured, and it has nonlinear elastic elements.
  Also, the joints are constructed by joint modules \cite{kawaharazuka2019musashi} which can measure joint angles for experimental evaluation.
  The joint angle values are a little noisy because of using analog potentiometers.
  Musashi has a biomimetic muscle arrangement, and each upper limb has 18 muscles (\#1--\#18).
  In this study, we mainly use five DOFs of the shoulder and elbow, and ten muscles (\#1--\#10), including one biarticular muscle (\#9), are involved with the DOFs.
  We represent these DOFs as S-p, S-r, S-y, E-p, and E-y (S means the shoulder, E means the elbow, and rpy means roll, pitch, and yaw).
  We basically control Musashi using length-based muscle stiffness control \cite{kawaharazuka2019longtime}.

  We will conduct four experiments using Musashi, shown in the classification of \figref{figure:reflex-classification}.
  As stretch reflex does not easily occur, we set $C^{stretch'}=15$ N which is larger than the maximum value of $\Delta{f}$ measured during ordinary motions, for all the subsequent experiments.
  Unless otherwise noted, we set $\Delta{l}^{stretch}=10.0$ and $\Delta{t}^{loose}=0.5$.
}%
{%
  本研究で用いる筋骨格ヒューマノイドMusashi \cite{kawaharazuka2019musashi}とその左腕の筋配置を\figref{figure:experimental-setup}に示す.
  Musashiは\secref{subsec:basic-structure}と同様に, 筋張力・筋長を測定でき, 非線形性要素を有している.
  また, 関節角度を測定可能な関節モジュール\cite{kawaharazuka2019musashi}によって関節が構成されており, 制御には用いないが, 実験の評価に用いることとする.
  人体を模倣した筋配置であり, 片腕で18の筋(\#1--\#18)を有し, 二関節筋も含まれる.
  本研究では主に肩と肘の5自由度を用い, それらに関係する筋は10本(\#1--\#10), うち二関節筋が1本(\#9)含まれている.
  これらの関節角度をS-p, S-r, S-y, E-p, E-yと表す(Sは肩関節, Eは肘関節, rpyはroll, pitch, yawを表す).

  本研究では以下において, \figref{figure:reflex-classification}に示された4つの動作を対象に, 実験を行っていく.
  なお, $C^{stretch'}$については, 伸張反射が容易に起こらないように, 基本的な動作を行った際に記録された$\Delta{f}$の最大値より大きな値である15 Nを全ての実験で採用した.
  また, 特に記載がない場合は$\Delta{l}^{stretch}=10.0$, $\Delta{t}^{loose}=0.5$とする.
}%

\begin{figure}[t]
  \centering
  \includegraphics[width=0.7\columnwidth]{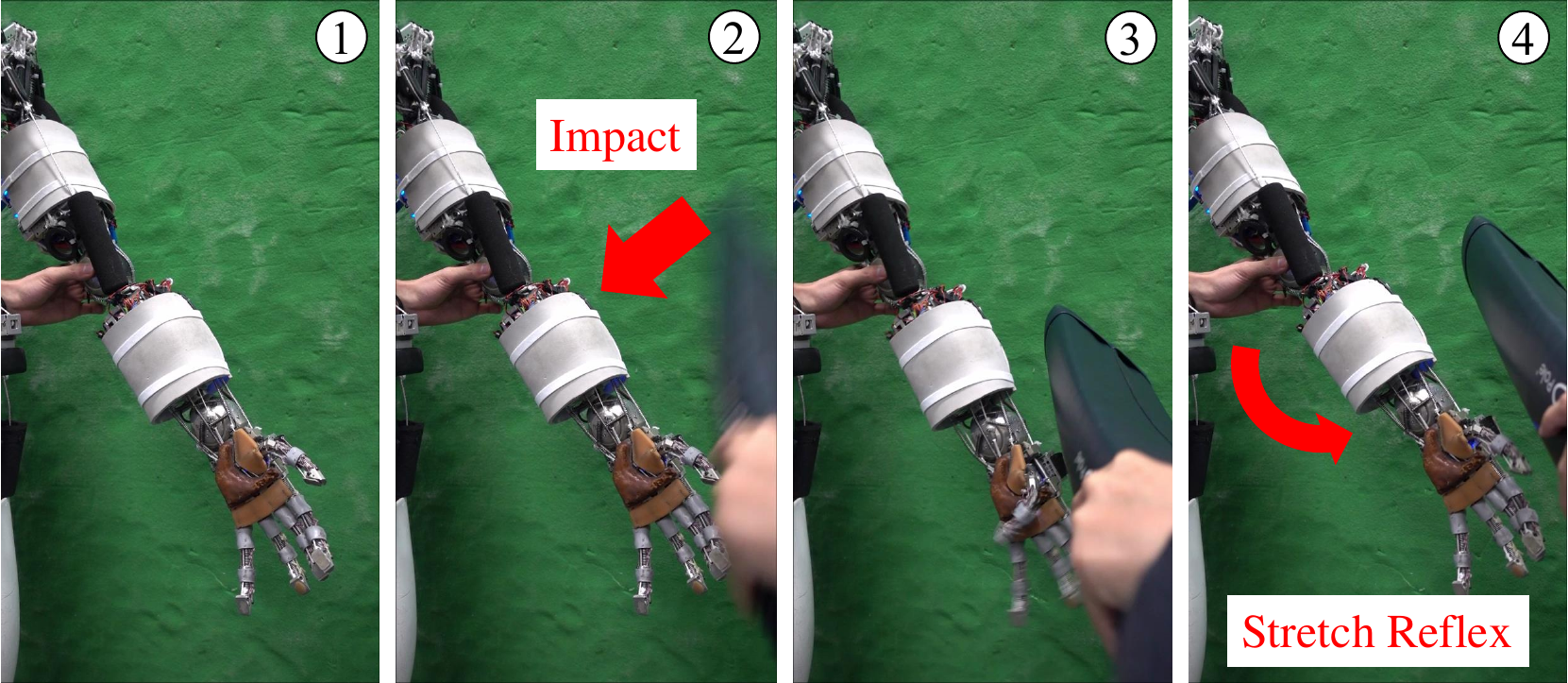}
  \caption{The experiment of the protective behavior of stretch reflex when adding sudden impact to the elbow joint.}
  \label{figure:protective-experiment}
  \vspace{-3.0ex}
\end{figure}

\begin{figure}[t]
  \centering
  \includegraphics[width=1.0\columnwidth]{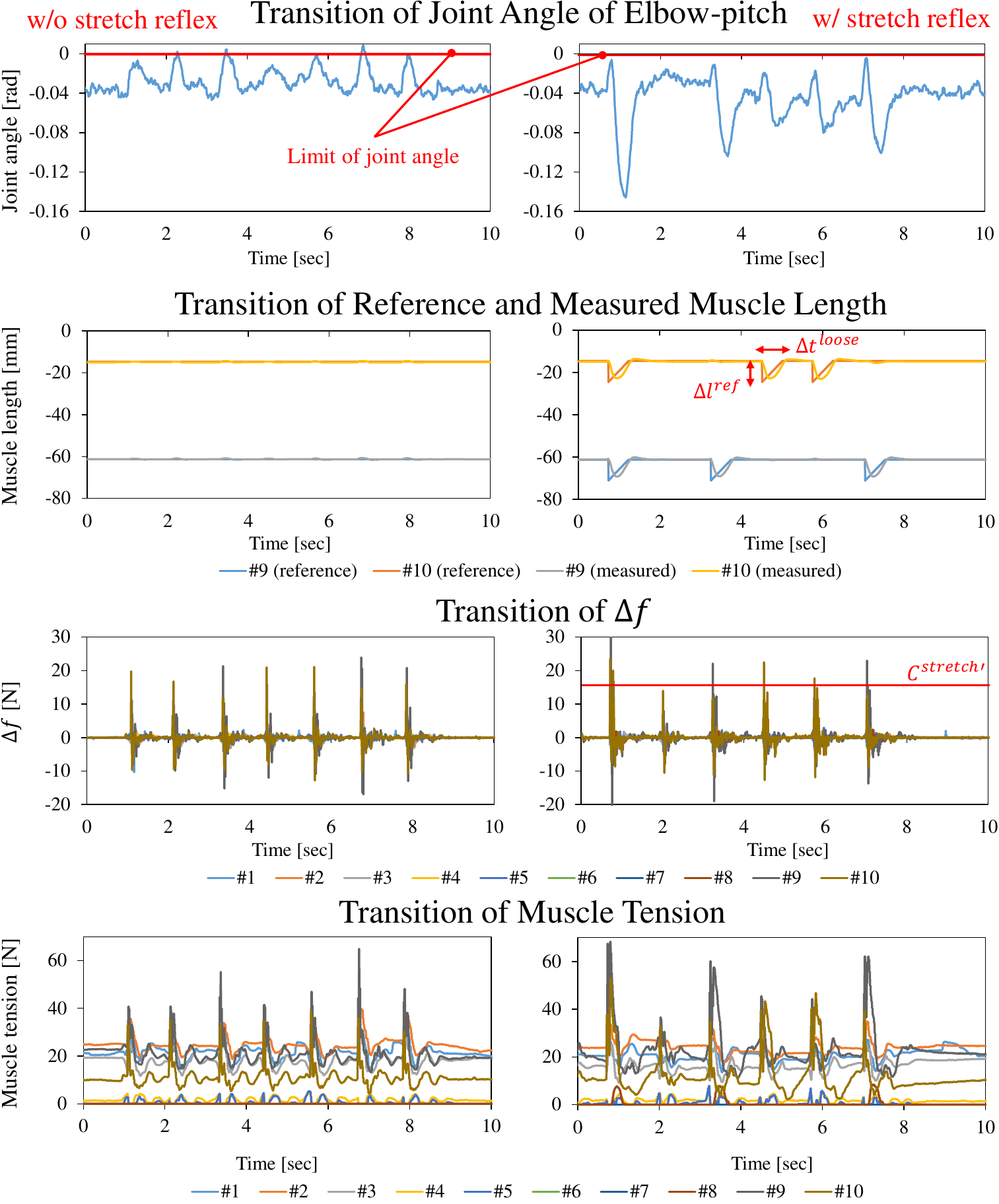}
  \caption{The experimental results of the protective behaviors with or without stretch reflex. The graphs show the joint angle of E-p, the comparison between the reference and measured muscle lengths regarding \#9 and \#10, the change in muscle tension $\Delta\bm{f}$, and muscle tension $\bm{f}$.}
  \label{figure:protective-graph}
  \vspace{-1.0ex}
\end{figure}

\subsection{Protective Application of Stretch Reflex} \label{subsec:protective-experiment}
\switchlanguage%
{%
  We conducted experiments regarding the protective behavior of stretch reflex.
  As shown in \figref{figure:protective-experiment}, E-p was bent to a small degree (about -0.04 rad), and impact forces were consecutively added to the forearm with the upper arm fixed.
  We verified the difference of behaviors with or without stretch reflex.

  We show the joint angle of E-p $\theta$, the comparison of reference $l^{ref}$ and measured muscle length $l$ of biceps brachii \#9 and brachialis \#10, which were mainly stretched by the impact, the changes in muscle tensions $\Delta\bm{f}$, and muscle tensions $\bm{f}$, in \figref{figure:protective-graph}.
  Without stretch reflex, when adding the impact force, the joint angle of E-p exceeded the joint angle limit of 0.0 rad.
  The part that physically limits the joint angle is made by a 3D printer, and a large force to deform it was exerted.
  On the other hand, with stretch reflex, the joint angle did not reach the joint angle limit.
  Stretch reflex occurred quickly when $\Delta{f}$ exceeded $C^{stretch'}$, the reference muscle length quickly contracted by $\Delta{l}^{stretch}$, and it loosened to the original length over $\Delta{t}^{loose}$.
  Stretch reflex occurred in each or both the biceps brachii \#9 and brachialis \#10.
  Although there was no large difference of muscle tension between with and without stretch reflex, large muscle tension continued a little longer with stretch reflex than without it.
}%
{%
  伸張反射を入れたときの危険回避に関する実験を行う.
  \figref{figure:protective-experiment}のように, E-pを少しだけ曲げ(約-0.04 rad), 上腕を掴んで固定した状態で, 前腕に衝撃を連続的に加える.
  このとき, 伸張反射を入れた時と入れない時における振る舞いの違いを検証した.

  E-pの関節角度$\theta$, 衝撃によって主に伸ばされる筋\#9と\#10の指令筋長$l^{ref}$と測定された筋長$l$の比較, 筋張力の変化$\Delta\bm{f}$, 筋張力$\bm{f}$の遷移を\figref{figure:protective-graph}に示す.
  伸張反射がない場合は, 衝撃を加えた際, 肘の関節角度限界である0.0 radに到達し, めり込んでいることがわかる.
  肘の関節角度制限部品が3Dプリンタであるため, それを変形させるような強い力が加わっている.
  これに対して, 伸張反射を入れた場合は, 関節角度限界にめり込むことはない.
  $\Delta{f}$が$C^{stretch'}$を超えた瞬間に伸張反射が起こり, 指令筋長が$-\Delta{l}^{stretch}$だけ一気に変化し, それが$\Delta{t}^{loose}$かけて元に戻っていくことがわかる.
  上腕二頭筋である\#9と上腕筋\#10は, 片方だけ伸張反射が起こったり, 同時に起こったりしている.
  筋張力は伸張反射がある場合とない場合では最大筋張力に大きな差はないが, 伸張反射を入れた場合は入れない場合に対して, 制御が入る分だけ筋張力が高い期間が少し長くなる.
}%

\begin{figure}[t]
  \centering
  \includegraphics[width=0.85\columnwidth]{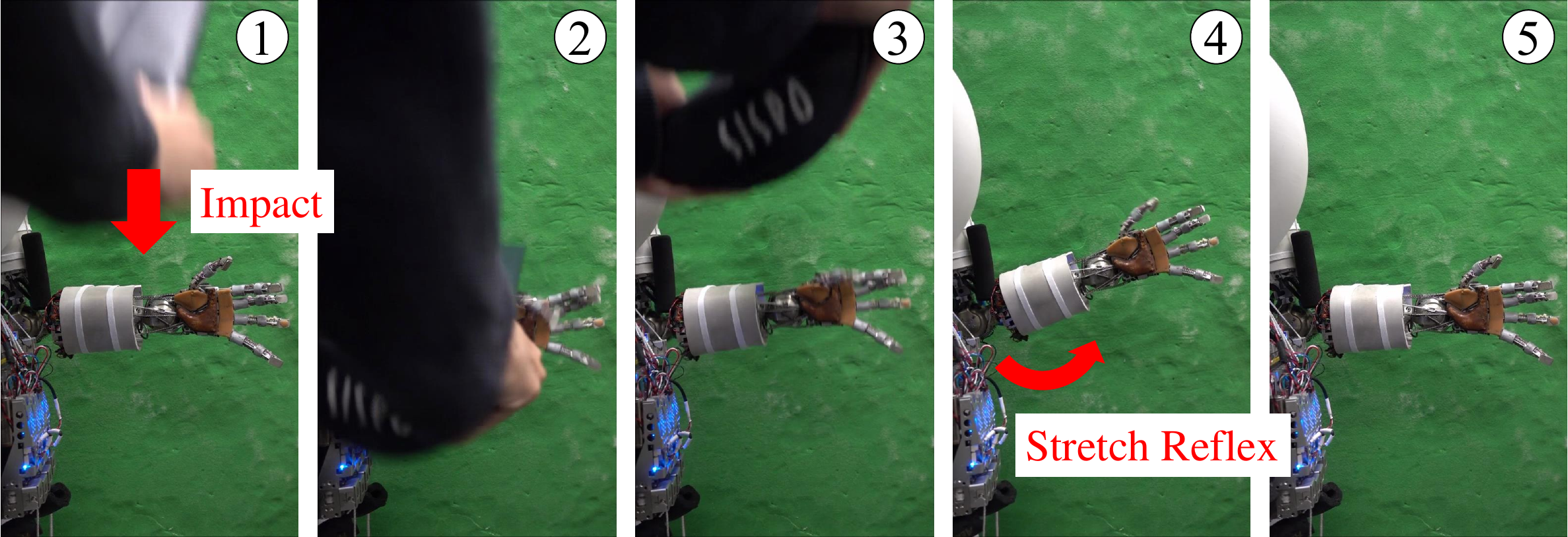}
  \caption{The experiment of stretch reflex for postural stability.}
  \label{figure:passive-experiment}
  \vspace{-3.0ex}
\end{figure}

\begin{figure*}[t]
  \centering
  \includegraphics[width=2.0\columnwidth]{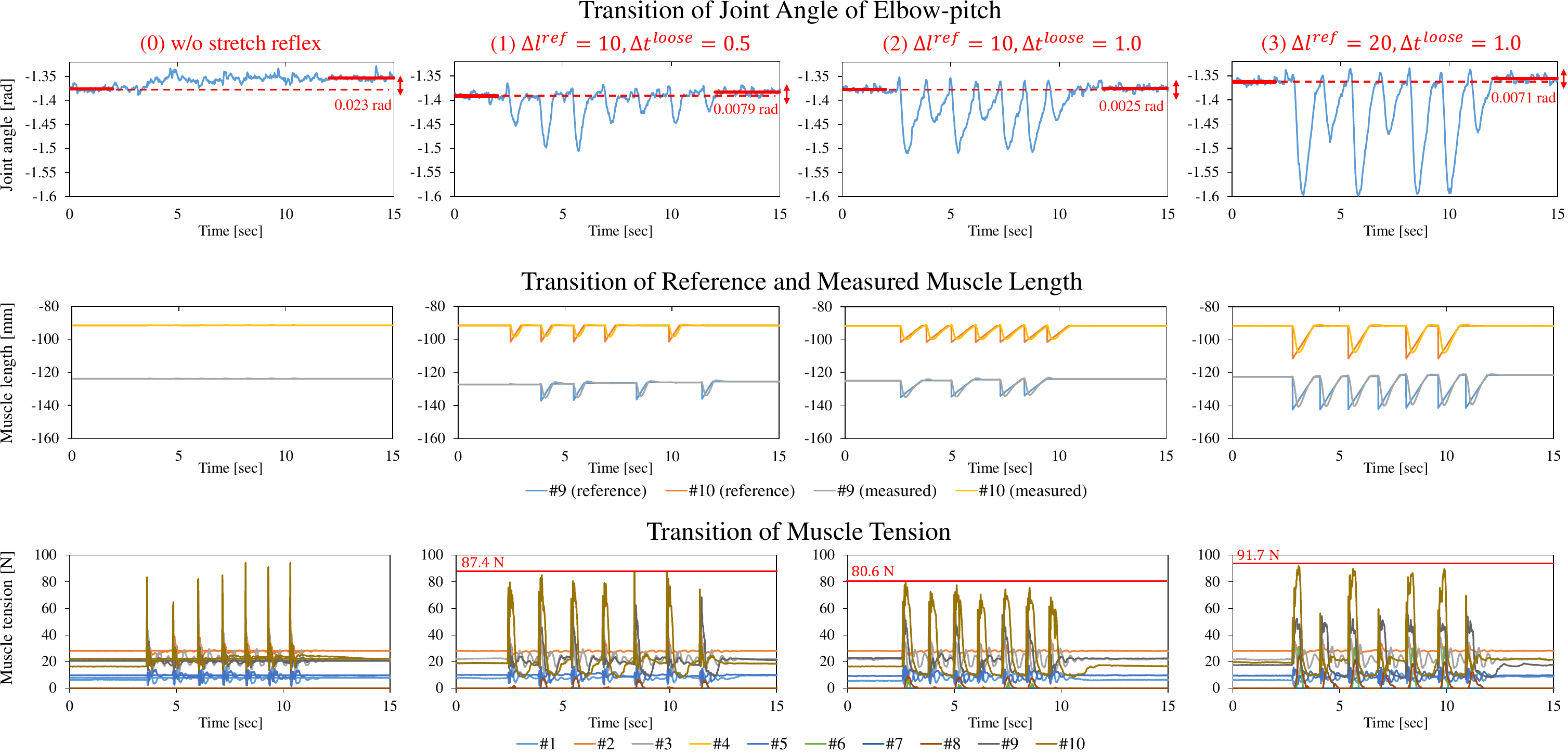}
  \caption{The experimental results of stretch reflex for postural stability. The graphs show the joint angle of E-p, the comparison between the reference and measured muscle lengths regarding \#9 and \#10, and muscle tension $\bm{f}$, without stretch reflex (0) and with stretch reflex: (1) $\Delta{l}^{stretch}=10, \Delta{t}^{loose}=0.5$, (2) $\Delta{l}^{stretch}=10, \Delta{t}^{loose}=1.0$, and (3) $\Delta{l}^{stretch}=20, \Delta{t}^{loose}=1.0$.}
  \label{figure:passive-graph}
  \vspace{-3.0ex}
\end{figure*}

\subsection{Application of Stretch Reflex for Postural Stability} \label{subsec:passive-experiment}
\switchlanguage%
{%
  We verified the effectiveness of stretch reflex for postural stability.
  As shown in \figref{figure:passive-experiment}, E-p was bent by about 90 deg, and impact force was consecutively added to the forearm.
  While \secref{subsec:protective-experiment} verified the protective behavior around the joint angle limit, we consider the change in posture before and after the impact force in this experiment.
  Also, we verify the difference in the behaviors by the change in parameters of stretch reflex.

  We show the joint angle of E-p $\theta$, the comparison of reference and measured muscle lengths of \#9 and \#10, and muscle tension $\bm{f}$, in \figref{figure:protective-graph}.
  We compared the cases without stretch reflex (0) and with stretch reflex of modified parameters: (1) $\Delta{l}^{stretch}=10, \Delta{t}^{loose}=0.5$, (2) $\Delta{l}^{stretch}=10, \Delta{t}^{loose}=1.0$, (3) $\Delta{l}^{stretch}=20, \Delta{t}^{loose}=1.0$.
  From the graphs of muscle lengths, according to the respective parameters, reasonable behaviors of reference muscle lengths were generated.
  Here, we verify the difference of joint angles of E-p between before and after the seven impacts.
  While the difference was 0.023 rad without stretch reflex, the differences with stretch reflex were (1) 0.0079 rad, (2) 0.0025 rad, and (3) 0.0071 rad.
  We can see that the change in joint angles by the sudden impact was inhibited by embedding stretch reflex.
  Also, although the maximum muscle tension was the largest regarding (3) with the largest $\Delta{l}^{stretch}$, a large difference could not be seen overall.
}%
{%
  伸張反射を入れた時の, 姿勢安定化に関する効果について検証を行う.
  \figref{figure:passive-experiment}のように, E-pを曲げ, その状態で前腕に連続的に衝撃を加える.
  \secref{subsec:protective-experiment}は関節角度限界付近で危険回避動作について検証したのに対して, 本実験では衝撃が加わる前と後での姿勢変化について考察する.
  また, 伸張反射のパラメータを変化させ, それによる振る舞いの違いについても検証する.

  E-pの関節角度$\theta$, \#9と\#10の指令筋長$l^{ref}$と測定された筋長$l$の比較, 筋張力$\bm{f}$の遷移を\figref{figure:protective-graph}に示す.
  伸張反射を入れない場合(0)と伸張反射を入れた場合: (1) $\Delta{l}^{stretch}=10, \Delta{t}^{loose}=0.5$, (2) $\Delta{l}^{stretch}=10, \Delta{t}^{loose}=1.0$, (3) $\Delta{l}^{stretch}=20, \Delta{t}^{loose}=1.0$について比較を行う.
  筋長のグラフから, それぞれのパラメータに従って, 伸張反射の際に妥当な筋長指令が生成されていることがわかる.
  このとき, それぞれ7回の衝撃を加える前と後における肘の関節角度の差分について検証する.
  伸張反射を入れないとき0.023 radなのに対して, (1)では0.0079 rad, (2)では0.0025 rad, (3)では0.0071 radであった.
  伸張反射を入れた場合は, 伸張反射を入れない場合に比べて, 衝撃による関節角度変化を抑制できていることがわかる.
  また, 最大筋張力については, $\Delta{l}^{stretch}$が大きな(3)について91.7 Nと最も大きかったが, 全体として大きな違いは見られなかった.
}%

\begin{figure}[t]
  \centering
  \includegraphics[width=0.8\columnwidth]{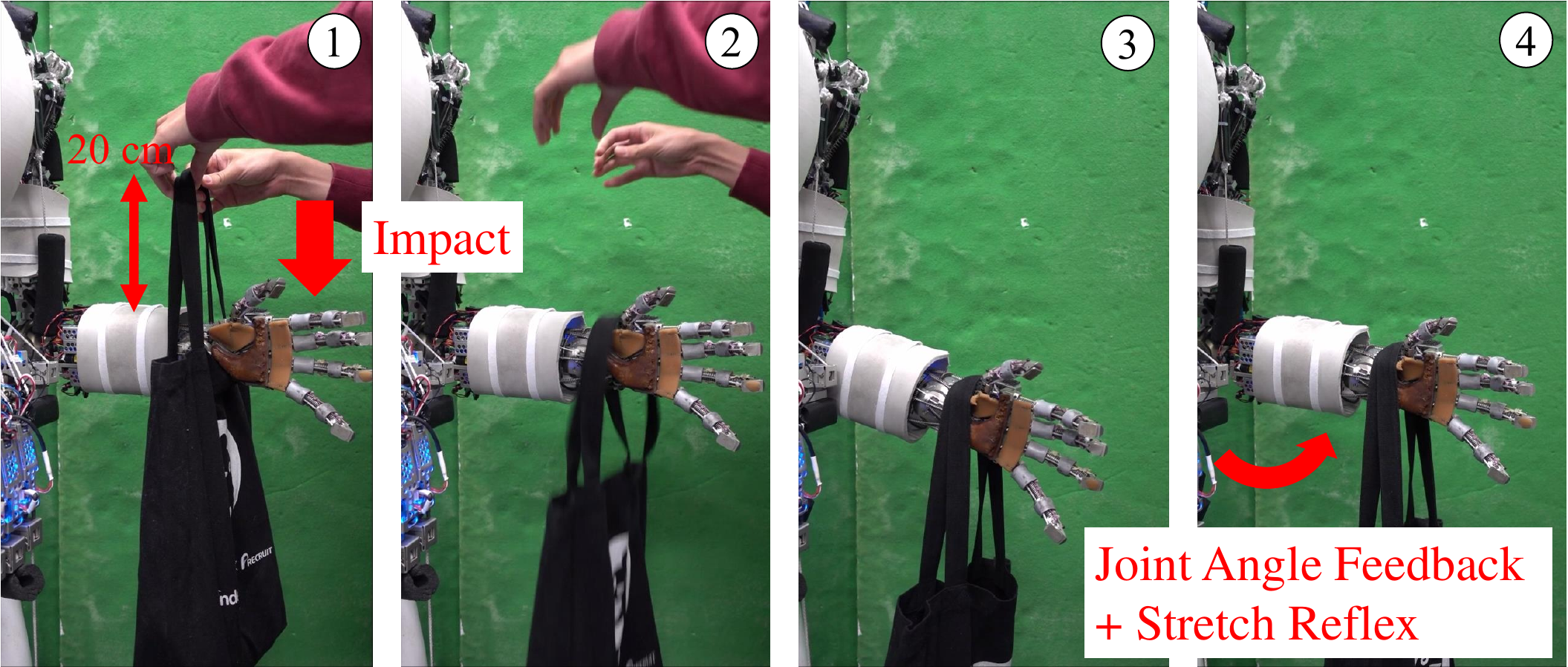}
  \caption{The experiment of using stretch reflex with joint angle feedback control for postural stability.}
  \label{figure:feedback-experiment}
  \vspace{-3.0ex}
\end{figure}

\begin{figure*}[t]
  \centering
  \includegraphics[width=2.0\columnwidth]{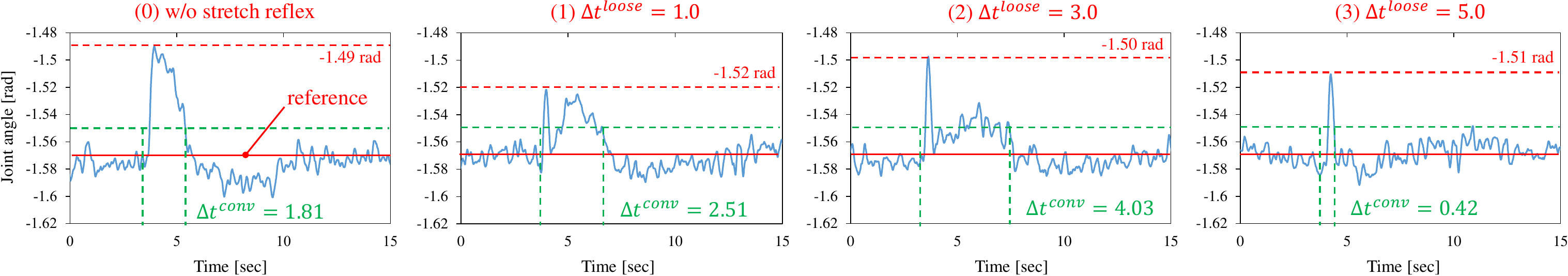}
  \caption{The experimental results of using stretch reflex with joint angle feedback control for postural stability. The graphs show the joint angle of E-p without stretch reflex (0) and with stretch reflex of modified parameters: (1) $\Delta{t}^{loose}=1.0$, (2) $\Delta{t}^{loose}=3.0$, and (3) $\Delta{t}^{loose}=5.0$.}
  \label{figure:feedback-graph}
  \vspace{-3.0ex}
\end{figure*}

\subsection{Application of Stretch Reflex with Joint Angle Feedback Control for Postural Stability} \label{subsec:feedback-experiment}
\switchlanguage%
{%
  We verified the effectiveness of stretch reflex for postural stability when adding joint angle feedback control.
  As shown in \figref{figure:feedback-experiment}, we conducted a feedback control to keep the joint angle of E-p at -90 deg, and dropped a bag with 3.6 kg dumbbell from about a 20 cm height to the forearm.
  We compared the behaviors by changing parameters as in \secref{subsec:passive-experiment}.
  The joint angle feedback control is a control that measures the current joint angle $\bm{\theta}$ and updates the virtual reference joint angle $\bm{\theta}^{virtual}$ by $\bm{\theta}^{virtual} \gets \bm{\theta}^{virtual} + \alpha(\bm{\theta}^{ref}-\bm{\theta})$.
  Here, $\bm{\theta}^{ref}$ is the reference joint angle (constantly -90 deg in this experiment).
  This feedback control is necessary because the muscles have hysteresis.
  After that, by using a mapping from joint angles to muscle lengths $\bm{h}$ \cite{kawaharazuka2018online}, we send reference muscle length $\bm{l}^{ref}$ as $\bm{l}=\bm{h}(\bm{\theta}^{virtual})$.
  In this study, we set $\alpha=0.3$, and this feedback control is executed at 5 Hz.

  We show the joint angle transition of E-p without stretch reflex (0) and with stretch reflex of modified parameters: (1) $\Delta{t}^{loose}=1.0$, (2) $\Delta{t}^{loose}=3.0$, and (3) $\Delta{t}^{loose}=5.0$, in \figref{figure:feedback-experiment}.
  The maximum joint angles when the impact force was added were (0) -1.49 rad, (1) -1.52 rad, (2) -1.50 rad, and (3) -1.51 rad, and the change in joint angle from the reference $\theta^{ref}=-1.57$ was the largest without stretch reflex.
  Thus, stretch reflex inhibited the maximum change in joint angle when the sudden impact was added.
  Also, we set a threshold $f^{thre}=-1.55$ [rad], and defined $\Delta{t}^{conv}$ as the time interval from the time of the impact to the earliest time at which $f$ does not exceed $f^{thre}$ from then.
  We can see that (0) $\Delta{t}^{conv}=1.81$ [sec], (1) $\Delta{t}^{conv}=2.51$ [sec], (2) $\Delta{t}^{conv}=4.03$ [sec], and (3) $\Delta{t}^{conv}=0.42$ [sec].
  While the convergence time when using stretch reflex with small $\Delta{t}^{loose}$ became longer than without stretch reflex, when $\Delta{t}^{loose}$ exceeds a certain threshold, the convergence time became much shorter than without stretch reflex.
}%
{%
  伸張反射に加えて, 関節角度フィードバッグを入れた時の姿勢安定化に関する効果について検証を行う.
  \figref{figure:feedback-experiment}のように, E-pの関節角度を-90度に保つようにフィードバッグ制御を加えた状態で, 3.6 kgのダンベルの入った袋を, 約20 cm上から前腕の先に対して落として荷重を増加させる実験を行う.
  本実験も前節と同様に, 伸張反射のパラメータを変更しながら比較する.
  ここで関節角度フィードバッグとは, 現在の関節角度$\bm{\theta}$を測定し, $\bm{\theta}^{ref}$を指令関節角度, $\bm{\theta}^{virtual}$を仮想的な指令関節角度として, $\bm{\theta}^{virtual} \gets \bm{\theta}^{virtual} + \alpha(\bm{\theta}^{ref}-\bm{\theta})$のように仮想的な指令関節角度を更新していく方法である.
  その後, 指令筋長$\bm{l}^{ref}$は関節角度と筋長の写像$\bm{h}$を用いて, $\bm{l}=\bm{h}(\bm{\theta}^{virtual})$ \cite{kawaharazuka2018online}のように計算されて実機に送られる.
  なお, 本研究では$\alpha=0.3$で, この制御は5 Hzで行われる.

  伸張反射を入れない場合(0)と伸張反射を入れた場合: (1) $\Delta{t}^{loose}=1.0$, (2) $\Delta{t}^{loose}=3.0$, (3) $\Delta{t}^{loose}=5.0$におけるE-pの関節角度の遷移を\figref{figure:feedback-experiment}に示す.
  このとき, 衝撃が加わった際の関節角度の最大値は(0) -1.49 rad, (1) -152 rad, (2) -1.50 rad, (3) -1.51 radと, 指令値$\theta^{ref}$である-1.57 radからの変化が伸張反射を入れない場合に最も大きかった.
  つまり, 伸張反射を入れた場合は衝撃が加わった際の最大関節角度変位が抑制されている.
  また, $f^{thre}=-1.55$ [rad]を閾値として, 衝撃が加わった瞬間からそれ以降$f$が$f^{thre}$を上回らない最初の時間までの期間を$\Delta{t}^{conv}$とする.
  このとき, (0) $\Delta{t}^{conv}=1.81$ [sec], (1) $\Delta{t}^{conv}=2.51$ [sec], (2) $\Delta{t}^{conv}=4.03$ [sec], (3) $\Delta{t}^{conv}=0.42$ [sec]となった.
  $\Delta{t}^{loose}$が小さいと, 伸張反射がない場合に比べて収束までの時間が長くなってしまうのに対して, ある閾値を超えると, 収束までの時間が伸張反射がない場合に比べて大幅に短くなることがわかった.
}%

\begin{figure}[t]
  \centering
  \includegraphics[width=0.8\columnwidth]{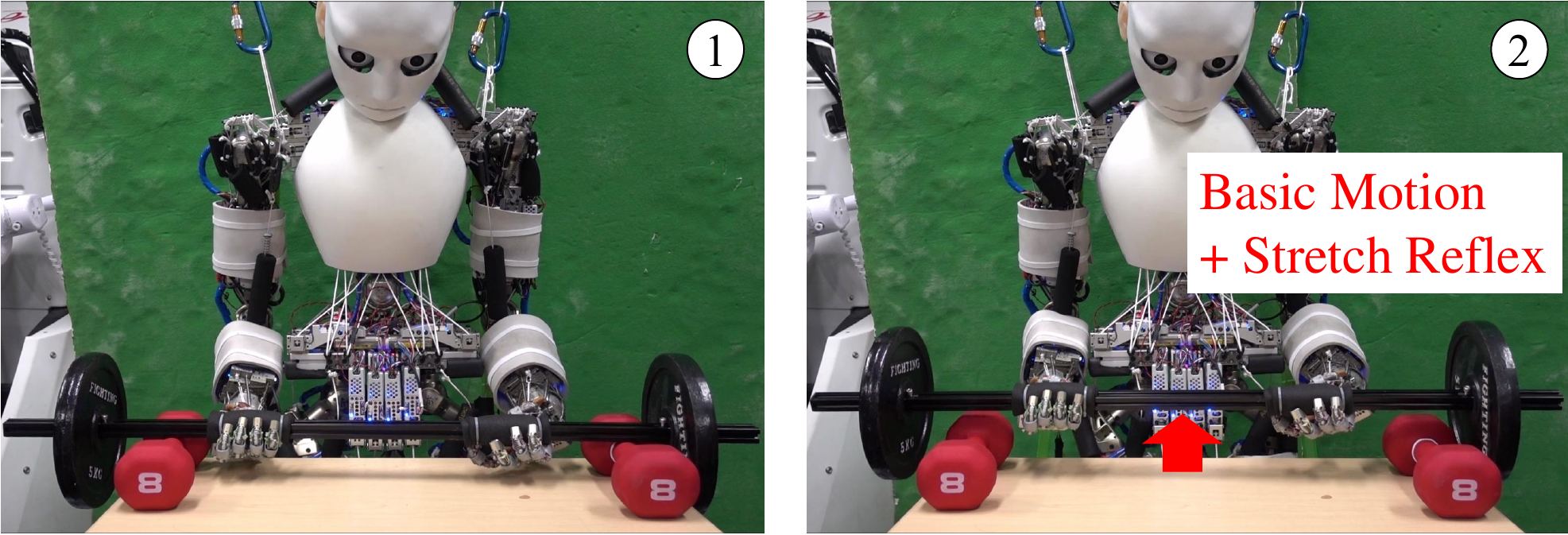}
  \caption{The experiment of actively using stretch reflex when lifting a heavy dumbbell.}
  \label{figure:active-experiment}
  \vspace{-1.0ex}
\end{figure}

\begin{figure}[t]
  \centering
  \includegraphics[width=1.0\columnwidth]{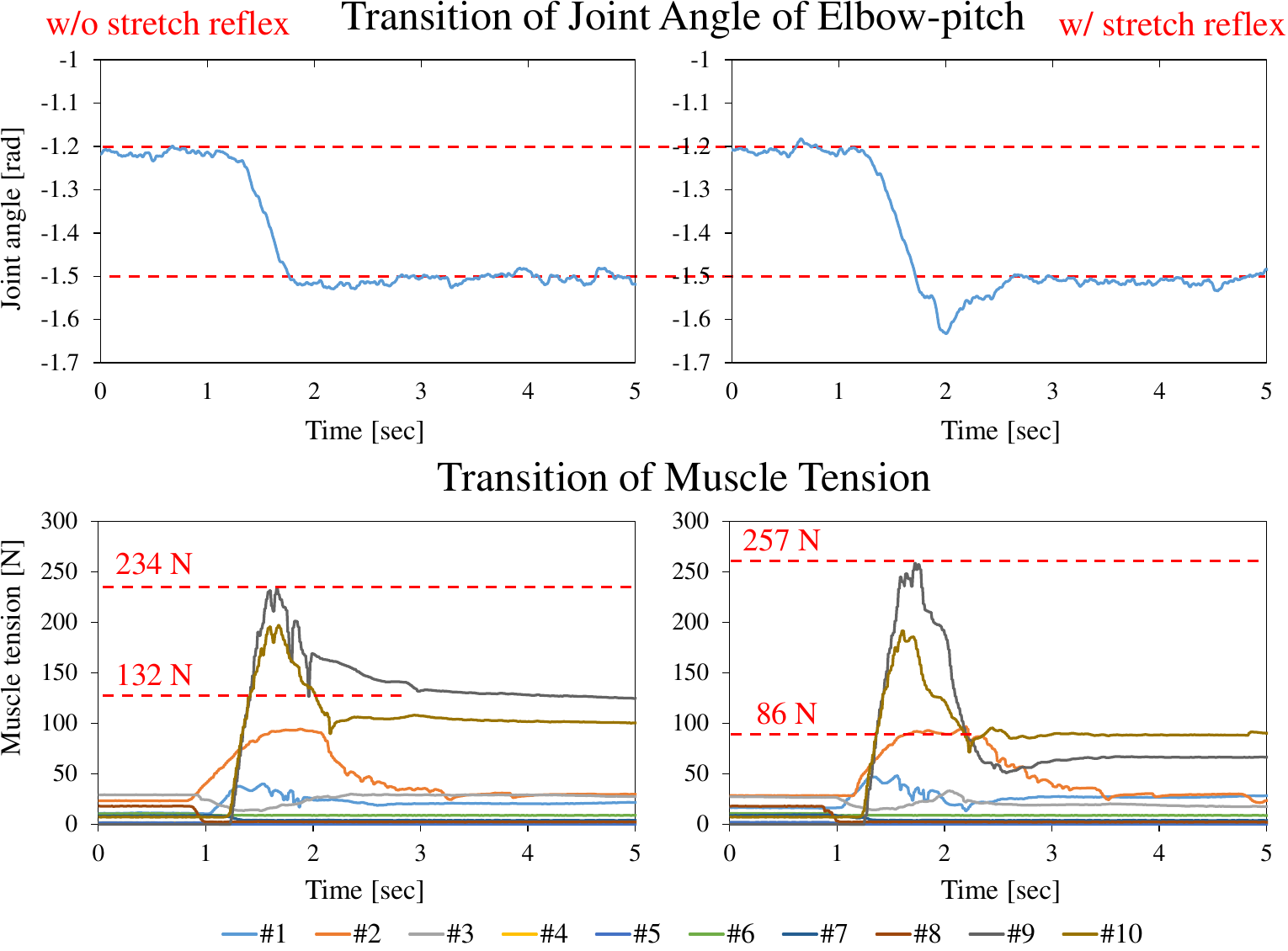}
  \caption{The experimental results of actively using stretch reflex when lifting a heavy dumbbell. The graphs show the joint angle of E-p and muscle tension, without stretch reflex (0) and with stretch reflex of modified parameters: (1) $\Delta{t}^{loose}=1.0$, (2) $\Delta{t}^{loose}=3.0$, and (3) $\Delta{t}^{loose}=5.0$.}
  \label{figure:active-graph}
  \vspace{-3.0ex}
\end{figure}

\subsection{Active Application of Stretch Reflex When Lifting a Heavy Dumbbell} \label{subsec:active-experiment}
\switchlanguage%
{%
  We verified the active applications of stretch reflex by taking an example of lifting a heavy dumbbell.
  As shown in \figref{figure:active-experiment}, we sent the motion of lifting a 10 kg dumbbell with both arms over one second.
  We verify the effectiveness of stretch reflex for the motion.

  We show the joint angle of E-p and muscle tension of the right arm with or without stretch reflex, in \figref{figure:active-graph}.
  The joint angles before and after lifting the dumbbell had no difference with or without stretch reflex.
  However, with stretch reflex, the joint angle largely changed in a moment and was reverted after that.
  The maximum muscle tensions were 234 N without stretch reflex and 257 N with stretch reflex.
  Also, the maximum muscle tensions after lifting the dumbbell were 132 N without stretch reflex and 86 N with stretch reflex.
  By embedding stretch reflex, the maximum muscle tension increased, but the muscle tension after lifting a heavy dumbbell decreased.
}%
{%
  伸張反射を能動的に用いる動作について, 重量物を持ち上げる動作を例に検証を行う.
  \figref{figure:active-experiment}のように, 重さ10 kgのダンベルを両手で持ち上げる動作を1秒で送る.
  このときの伸張反射の効果を検証する.

  伸張反射を入れない場合と入れた場合について, 右腕のE-pの関節角度と筋張力の遷移を\figref{figure:active-graph}に示す.
  伸張反射を入れない場合と入れた場合において, ダンベルを持ち上げる前と後の関節角度変化はほとんど変わらない.
  しかし, 伸張反射を入れた場合は, 一瞬だけ関節角度が大きく変化し, その後戻っていくような推移が見られる.
  筋張力は, 伸張反射を入れない場合は最大で234 N, 入れた場合は最大で257 Nとなった.
  また, 持ち上げた後の筋張力は, 伸張反射を入れない場合は132 N, 入れた場合は86 Nとなった.
  伸張反射を入れることで, 一瞬の最大筋張力は増大するものの, 持ち上げたあとの筋張力は減少することが分かった.
}%

\section{Discussion} \label{sec:discussion}
\switchlanguage%
{%
  We summarize and discuss the four experimental results.

  In \secref{subsec:protective-experiment}, we verified the behaviors of stretch reflex around the joint angle limit when sudden impact is added.
  While the joint angle reaches the limit and large force is applied to the part that physically limits the joint angle without stretch reflex, the protective behavior can be seen with stretch reflex, as the joint angle does not reach the limit.

  In \secref{subsec:passive-experiment}, we verified the effectiveness of stretch reflex for postural stability by bending the elbow and adding the sudden impact while changing parameters.
  With stretch reflex, the change in joint angles before and after the sudden impact decreases compared to without stretch reflex.
  Because the muscles of the musculoskeletal humanoid have hysteresis caused by friction, the joint angle gradually changes by the impact force without stretch reflex.
  On the other hand, by embedding stretch reflex, the joint angle changed by the impact force is reverted, and the same joint angle can be constantly kept.
  While the muscle tension decreases by decreasing $\Delta{l}^{stretch}$, the effect of stretch reflex becomes weak.
  While the robot can quickly respond to the next impact by decreasing $\Delta{t}^{loose}$, the effect of stretch reflex becomes weak if $\Delta{t}^{loose}$ is too short, and $\Delta{t}^{loose}$ should be long to a certain degree in terms of feedback control explained subsequently.
  Thus, there is a tradeoff of parameters of stretch reflex.

  In \secref{subsec:feedback-experiment}, we verified the effectiveness of stretch reflex with joint angle feedback control for postural stability when suddenly receiving a heavy object.
  With stretch reflex, the maximum change in joint angle when the sudden impact is added is inhibited compared to without stretch reflex.
  Also, when $\Delta{t}^{loose}$ is short, the muscle length is quickly reverted after stretch reflex, the joint angle feedback control is executed as without stretch reflex, and the convergence time becomes longer.
  On the other hand, when $\Delta{t}^{loose}$ is sufficiently long, after the joint angle is suddenly reverted to $\theta^{ref}$ by stretch reflex, $\theta^{ref}$ can be constantly kept with the joint angle feedback control.
  The longer $\Delta{t}^{loose}$ is, the higher the contribution of the joint angle feedback becomes compared with stretch reflex.
  Thus, by setting $\Delta{t}^{loose}$ appropriately, quick postural stability after the impact or burden is enabled.

  In \secref{subsec:active-experiment}, as an example of inducing stretch reflex on purpose, we verified the effectiveness of stretch reflex when lifting a heavy object.
  With stretch reflex, while the muscle tension increases for a moment, the final muscle tension decreased.
  By adding excessive force to the heavy object, stretch reflex is induced and muscle tension suddenly increases.
  Due to the hysteresis of muscles caused by friction, while large muscle tension is necessary to exceed the friction without stretch reflex, by embedding stretch reflex, the muscle quickly moves for a moment, the friction is exceeded, and the final muscle tension is reduced.
  Thus, by adding large force to the environment and inducing stretch reflex on purpose, the robot can realize the task by reducing the final muscle tension.

  From these experiments, stretch reflex avoids the joint angle limit and enables the fast convergence of joint angles when combined with joint angle feedback control.
  Also, stretch reflex overcomes the hysteresis of muscles, inhibits the change of posture by sudden impact, and contributes to the reduction of muscle tension by making use of it actively.
  While previous studies have discussed stretch reflex mainly in simulation or for postural stability of lower limbs, these results can be obtained for the first time when the characteristics of the actual robot (e.g. friction) are included.

  In this study, while we consider the difference of behaviors by the change of parameters, we need to automatically determine the parameters for desired behaviors.
  The parameters of human stretch reflex are known to be changed by the task or environment \cite{doemges1992modulation}, and we need to embed this mechanism into the musculoskeletal humanoid.
}%
{%
  これまでの4つの実験の結果をまとめ, 考察する.

  \secref{subsec:protective-experiment}では, 伸張反射がない場合とある場合について, 関節角度限界付近において衝撃を加えたときの挙動を確認した.
  伸張反射がない場合は関節角度限界に達し, 関節のリミット部品に対して大きな力がかかってしまう一方, 伸張反射を入れた場合は関節角度限界に達さないような危険回避を行うことができる.

  \secref{subsec:passive-experiment}では, 肘を大きく曲げた状態で衝撃を加え, 伸張反射のパラメータを変化させながら, そのときの姿勢安定化の効果について確認した.
  伸張反射を入れることで, 入れない場合に比べて, 衝撃を加える前と後での関節角度の変化が小さくなた.
  筋骨格ヒューマノイドには摩擦によるヒステリシスが大きく, 伸張反射がない場合は, 衝撃によって徐々に関節角度が変化してしまう.
  これに対して, 伸張反射を入れることで一度衝撃で変化した関節角度が元に戻り, 常に衝撃前と同じ関節角度を保持し続けることができる.
  $\Delta{l}^{stretch}$を小さくすればかかる筋張力を小さくできる一方伸張反射の効果は薄まり, $\Delta{t}^{loose}$を小さくすれば伸張反射の期間を短くしてすぐに次の衝撃に対応できる一方, 小さくし過ぎると伸張反射の効果が薄れ, 後述するフィードバッグ等の観点からもある程度の長さがあった方が良い.
  このように, 伸張反射のパラメータにはトレードオフが存在する.

  \secref{subsec:feedback-experiment}では, 重い物体を突然持たせたときの, 関節角度フィードバッグと伸張反射を同時に入れた場合の姿勢安定化について確認した.
  伸張反射を入れることで, 入れる前に比べて衝撃時の最大関節角度変化が抑制された.
  また, $\Delta{t}^{loose}$が小さいと, 伸張反射後にすぐに筋が戻ってしまい, 伸張反射がない場合と同じようにフィードバッグがかかり, より収束までの時間が遅くなってしまう.
  それに対して, 十分な長さの$\Delta{t}^{loose}$が確保されると, 衝撃後に伸張反射で$\theta^{ref}$まで一気に戻った後, そのまま関節角度フィードバッグ制御と相まって$\theta^{ref}$が保ち続けられるような現象が見られた.
  つまり, $\Delta{t}^{loose}$を適切に設定することで, 衝撃や荷重が加わった後の素早い姿勢安定化が可能となるのである.

  \secref{subsec:active-experiment}では, 能動的に伸張反射を起こす例として, 重量物の持ち上げの際の伸張反射の効果について確認した.
  伸張反射を入れることで, 一瞬だけ最大筋張力が増大してしまうものの, 最終的な筋張力は減少した.
  非常に重い物体に無理な力をかけることで伸張反射が起き, 一瞬だけ筋張力が増大する.
  摩擦による筋のヒステリシスが大きいため, 通常の動作時にはその摩擦を乗り越えるために大きな筋張力が必要となるのに対して, 伸張反射を入れた場合は一瞬だけ筋を速い速度で動かすため, その摩擦を乗り越えて最終的に収束する筋張力が減少することがわかった.
  つまり, わざと無理な力を環境にかけることで伸張反射を誘発させ, それを利用することで最終的な必要筋張力を削減してタスクを遂行することができる.

  このように, 伸張反射は関節角度限界を回避し, 関節フィードバックと合わせることで速い収束を促すことができる.
  また, 筋の摩擦によるヒステリシスを克服し, 衝撃による姿勢変化を抑制し, 能動的に使用することで筋張力削減に寄与する.
  これまで主にシミュレーションや下肢における姿勢安定化のみについて議論されてきたのに対して, これらの結果は実機における摩擦等の影響が加味されて始めて得られるものである.

  本研究ではパラメータの変化による挙動の違いについて見たが, これらを自動的に決定する手法が必要である.
  人間の伸張反射のパラメータは, タスクや環境によって変化することが知られており\cite{doemges1992modulation}, これの機構を筋骨格ヒューマノイドにも組み込むことが求められる.
}%

\section{CONCLUSION} \label{sec:conclusion}
\switchlanguage%
{%
  In this study, we embedded stretch reflex into the upper limb of the actual musculoskeletal humanoid and verified its effectiveness.
  We classified the applications of stretch reflex into passive and active.
  Regarding passive applications, we handled the protective behavior, and postural stability with or without joint angle feedback control, when sudden impact is added.
  Regarding active applications, we handled the motion of lifting a heavy object by inducing stretch reflex on purpose.
  By embedding stretch reflex, the robot can cope with the sudden impact around the joint angle limit and inhibit the burden to the joint.
  Also, when the sudden impact is added, stretch reflex can inhibit the change in joint angles caused by hysteresis, and realize the fast convergence of joint angles with joint angle feedback control.
  When lifting a heavy object, by making use of stretch reflex with excessive force added to the environment, the robot can finally realize the task with smaller muscle tensions.
  Thus, we succeeded in discovering multiple effective applications of human stretch reflex for the upper limb of the actual musculoskeletal humanoid.

  In future works, we would like to verify the effectiveness when combining this study with goldi tendon reflex and reciprocal innervation, and conduct more human-like motions.
}%
{%
  本研究では, 筋骨格ヒューマノイドの上肢実機に伸張反射を組み込み, その有用性について検証した.
  受動的と能動的に分け, 受動的動作については衝撃が加わった際の危険回避・姿勢保持・関節フィードバッグ制御を入れた際の姿勢保持について, 能動的動作については重量物体の持ち上げ動作について有用性を検証した.
  伸張反射を入れることで, 関節角度限界近くでの衝撃に対応し, 関節に負荷をかけないようにすることができる.
  また, 衝撃を加えられた際に, ヒステリシスによる関節角度変化を抑制し, パラメータを調整することで, フィードバッグ制御と合わせて素早い収束を実現することができる.
  重量物体を持ち上げる際は, わざと身体に衝撃を加えることによって誘発される伸張反射を利用することで, 最終的により小さな筋張力でタスクを遂行することができる.
  つまり, 筋骨格ヒューマノイドの上肢において, 実験から人間の伸張反射の複数の有用なアプリケーションを見出すことに成功した.

  今後は, そのゴルジ腱反射や相反性神経支配等と合わた際の有用性の検証, より人間らしい動作を行っていきたい.
}%

{
  \bibliographystyle{IEEEtran}
  \bibliography{main}
}

\end{document}